\documentclass{article}

\usepackage[preprint]{neurips_2024}

\usepackage[utf8]{inputenc} % allow utf-8 input
\usepackage[T1]{fontenc}    % use 8-bit T1 fonts
\usepackage{hyperref}       % hyperlinks
\usepackage{url}            % simple URL typesetting
\usepackage{booktabs}       % professional-quality tables
\usepackage{amsfonts}       % blackboard math symbols
\usepackage{nicefrac}       % compact symbols for 1/2, etc.
\usepackage{microtype}      % microtypography
\usepackage{xcolor}         % colors

\usepackage{kotex}
\usepackage{booktabs} % for professional tables
\usepackage{subcaption}
\usepackage{wrapfig}
\usepackage{enumitem}
\usepackage[hang, flushmargin]{footmisc}
\usepackage{hyperref}

\usepackage{multirow}
\usepackage{graphicx}
\usepackage[normalem]{ulem} % Ensure this is included for \uline command
\useunder{\uline}{\ul}{} % Ensure this is included for \ul command
\usepackage{adjustbox}
\usepackage{array}

\title{EXAONE Path 2.0: Pathology Foundation Model \\ with End-to-End Supervision}
\author{%
  LG AI Research\thanks{\textbf{Core contributors}: Myungjang Pyeon, Janghyeon Lee, Minsoo Lee, Juseung Yun, Hwanil Choi, Jonghyun Kim, Jiwon Kim, Yi Hu, Jongseong Jang, Soonyoung Lee, \textbf{Contributors}: Yeonuk Jeong, Hyunjoo Yeo, Yong Min Park, Edward Lee, Woohyung Lim}
}

\begin{document}

\maketitle

% \vspace{-20pt}
\begin{abstract}
In digital pathology, whole-slide images (WSIs) are often difficult to handle due to their gigapixel scale, so most approaches train patch encoders via self-supervised learning (SSL) and then aggregate the patch-level embeddings via multiple instance learning (MIL) or slide encoders for downstream tasks.
However, patch-level SSL may overlook complex domain-specific features that are essential for biomarker prediction, such as mutation status and molecular characteristics, as SSL methods rely only on basic augmentations selected for natural image domains on small patch-level area.
Moreover, SSL methods remain less data efficient than fully supervised approaches, requiring extensive computational resources and datasets to achieve competitive performance.
To address these limitations, we present EXAONE Path 2.0, a pathology foundation model that learns patch-level representations under direct slide-level supervision.
Using only 37k WSIs for training, EXAONE Path 2.0 achieves state-of-the-art average performance across 10 biomarker prediction tasks, demonstrating remarkable data efficiency. 
% The EXAONE Path 2.0 model weight is available at \texttt{https://huggingface.co/LGAI-EXAONE/EXAONE-Path-2.0}.
\end{abstract}

\begin{figure}[h]
    \centering
    \begin{subfigure}[h]{0.5\textwidth}
        \centering
        \includegraphics[width=\textwidth,height=0.8\textheight,keepaspectratio]{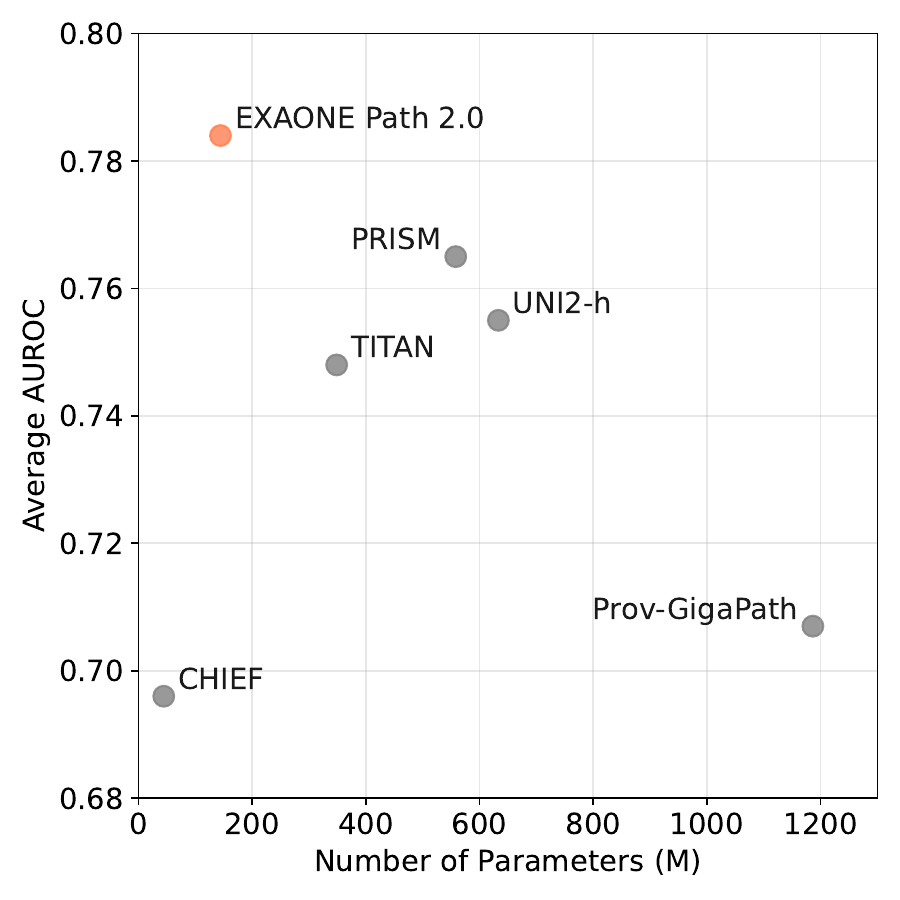}
        \caption{Model Size vs. Average AUROC}
        \label{fig:params_vs_auroc}
    \end{subfigure}%
    \hfill%
    \begin{subfigure}[h]{0.5\textwidth}
        \centering
        \includegraphics[width=\textwidth,height=0.8\textheight,keepaspectratio]{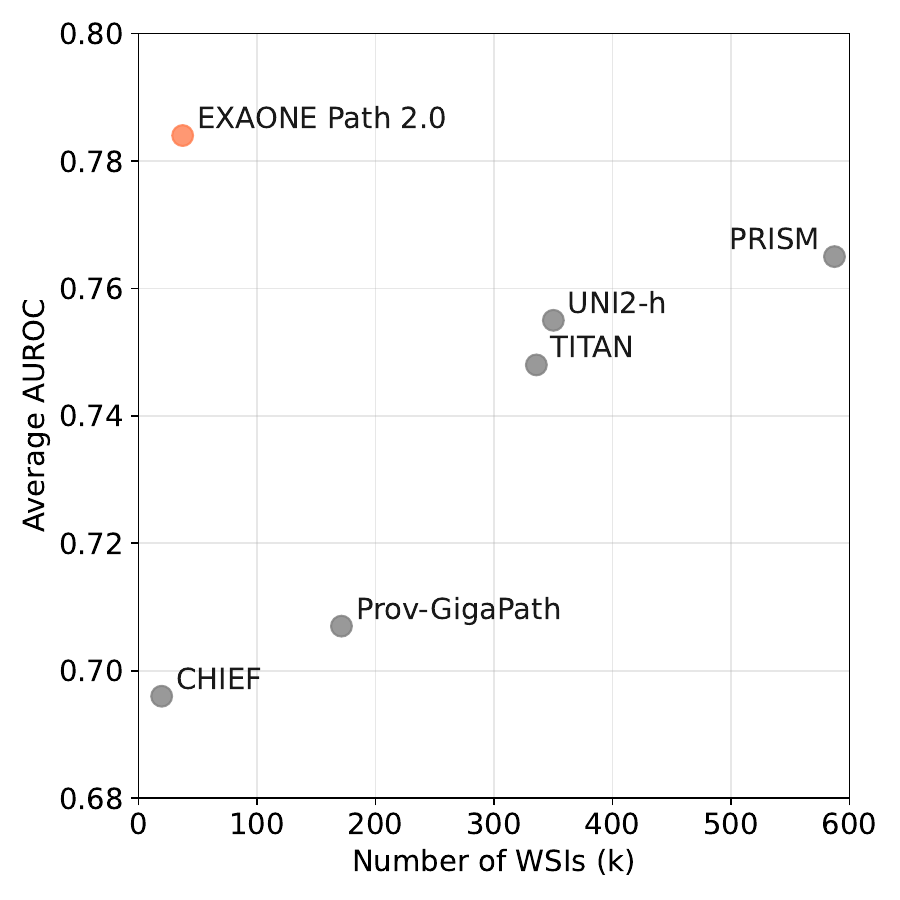}
        \caption{Training Data Size vs. Average AUROC}
        \label{fig:wsis_vs_auroc}
    \end{subfigure}
    \caption{Performance comparison of models based on the number of parameters and the number of WSIs used for training. The average AUROC is obtained by averaging AUROC scores on 10 biomarker prediction tasks. Notably, EXAONE Path 2.0 achieves high performance despite having fewer parameters and using fewer WSIs compared to other models, demonstrating its efficiency.}
    \label{fig:model_comparison}
\end{figure}

% \section*{\blue{*Important Notice*}}
% \blue{This document is valid only for \name{} 2.0. Also, please refer to the license at the end of the document, when using this model.}

\section{Introduction}

Digital pathology has emerged as a critical domain for AI-driven healthcare applications, with whole-slide images (WSIs) presenting unique computational challenges due to their gigapixel scale \cite{hipt, prism, gigapath}. Current approaches typically follow a two-stage paradigm: training patch-level encoders through self-supervised learning methods such as DINO \cite{dino} and DINOv2 \cite{dinov2}, then aggregating patch-level embeddings using multiple-instance learning (MIL) or slide-level encoders for downstream prediction tasks \cite{titan, clam, prism, gigapath}.

Although this paradigm has shown promise, it has fundamental limitations in the digital pathology field. Self-supervised patch-level pretraining does not guarantee to capture complex domain-specific features that are essential for biomarker prediction, such as mutation status or other molecular characteristics, as self-supervised learning (SSL) methods rely only on basic augmentations selected for natural image domains on small patch-level area. Moreover, these approaches demonstrate inferior data efficiency compared to fully supervised methods, requiring extensive computational resources and large datasets to achieve competitive performance \cite{koppula2022should, wang2021solving}.

To address these limitations, we introduce EXAONE Path 2.0, a pathology foundation model that learns patch-level representations under direct slide-level supervision. Our approach fundamentally differs from existing methods by incorporating multiple slide-level labels during patch encoder training, enabling the model to learn clinically relevant features more effectively.

Our results demonstrate that EXAONE Path 2.0 achieves superior average performance across all evaluated tasks while requiring substantially fewer training samples than competing methods, marking a significant advancement in computational pathology.

% \section{Related Work}
% \input{2.related_work}

\section{Modeling}

\begin{figure}
    \centering
    \vspace{-10pt}
    \includegraphics[width=0.98\linewidth]{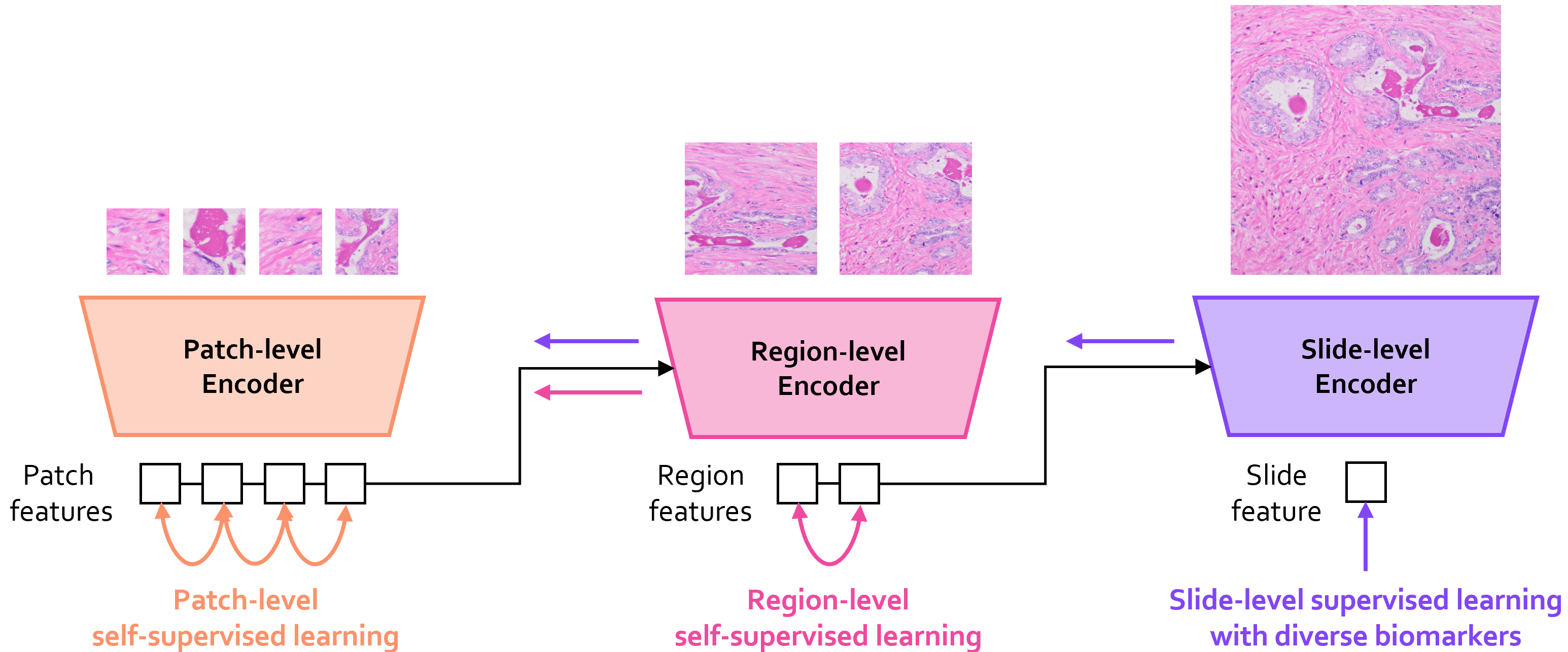}
    \caption{End-to-end hierarchical learning in EXAONE Path 2.0. Slide-level supervised signals propagate through all three hierarchical ViT stages, enabling end-to-end learning of clinically relevant representations from patch to slide level. Self-supervised learning at patch and region levels enhances feature robustness and leverages unlabeled data.}
    \label{fig:concept}
\end{figure}

\subsection{Overcoming the Prohibitive Computational Costs of Gigapixel Image Training}
% HIPT 구조를 따라서 3단계 ViT로 구성
% curriculum learning 적용 -- 1단계/2단계 ViT를 DINO로 학습 (256, 1024 res), 1단계 DINO + sup loss
% sup loss 계산할때 activation checkpointing과 cpu offloading으로 한번에 메모리에 다 올리지 않음

Training on gigapixel whole-slide images presents significant computational challenges due to memory constraints and processing requirements. To address these limitations, we employ a combination of hierarchical architecture design, curriculum learning, and efficient memory management techniques.

\textbf{Architecture Design.} We adopt a three-stage Hierarchical Image Pyramid Transformer (HIPT) \cite{hipt} architecture. This hierarchical design reduces computational complexity by processing patches at progressively higher levels of abstraction rather than directly processing gigapixel images at full resolution, enabling more efficient handling of large-scale WSIs. The first-stage ViT processes individual patches, the second-stage ViT aggregates patch-level features into region-level representations, and the third-stage ViT processes the entire slide by integrating all region-level features.

\textbf{Curriculum Learning.} To manage the computational burden of end-to-end training across all stages simultaneously, we implement a two-stage curriculum learning approach with progressive resolution scaling. In the first curriculum stage, we apply 256×256 DINO loss to the first-stage ViT and 1024×1024 DINO loss to the second-stage ViT, establishing hierarchical visual representations without requiring full three-stage end-to-end computation. In the next curriculum stage, we continue applying 256×256 DINO loss to the first-stage ViT while scaling up to 4096×4096 regions for the second-stage ViT, and introduce slide-level supervised cross-entropy loss to propagate gradients into the entire three-stage model processing the full slide. This curriculum approach significantly reduces computational overhead by avoiding the need to process all stages at maximum resolution during every training iteration.

\textbf{Memory Management.} To further manage the computational demands of processing entire WSIs, we employ activation checkpointing and CPU offloading strategies. Rather than loading all patch embeddings into GPU memory at once, we dynamically compute and transfer activations as needed during supervised loss calculation. This approach significantly reduces memory requirements while maintaining training efficiency, enabling us to process gigapixel images with limited computational resources.

\subsection{Learning Generalizable Representations across Multiple Biomarker Prediction Tasks}
% multi-task learning: cancer subtyping 
% Small data, deep network 상황에서 overthinking을 줄이기 위해 1st stg model과 CLAM을 활용해 downstream task에 적용

To learn representations that generalize across diverse biomarker prediction tasks while maintaining computational efficiency, we employ a multi-task learning framework combined with an early exit strategy for downstream task adaptation.

\textbf{Multi-Task Learning Framework.} We implement a multi-task learning approach that jointly optimizes across multiple complementary objectives. Our training encompasses three primary categories of tasks: (1) cancer subtyping across 33 cancer types, (2) tissue type classification across 12 organ systems, and (3) molecular biomarker prediction including pan-cancer and cancer-specific mutation status, microsatellite instability, and hormone receptor subtyping. This multi-task learning strategy jointly optimizes for these diverse prediction objectives, encouraging the model to learn shared representations that capture fundamental pathological patterns across different scales of biological organization. The joint optimization helps prevent overfitting to individual tasks while improving generalization across the entire spectrum of downstream applications.

\textbf{Early Exit Strategy for Downstream Adaptation.} To further mitigate overfitting in the small data and deep network regime, we adopt a shallow network approach that leverages early representations rather than the full hierarchical model \cite{shallow-deep-net}. Specifically, we leverage the first-stage model in combination with Clustering-constrained Attention Multiple Instance Learning (CLAM) \cite{clam} for downstream task adaptation. Rather than fine-tuning the entire hierarchical network, this early exit approach uses the robust patch-level features from the first-stage model, while CLAM efficiently aggregates these features for slide-level predictions. This strategy significantly reduces computational overhead during downstream task adaptation while avoiding the pitfalls of overfitting commonly observed in pathology applications with limited data.

\section{Experiments}
% In this section, we describe the experimental setup used to evaluate our proposed model.
% We first introduce the baseline models used for comparison, followed by a description of the evaluation protocols.
% We then present the 10 slide-level benchmark tasks curated from both private and public datasets to comprehensively assess model performance.

\subsection{Training Data}
EXAONE Path 2.0 is trained on 37,195 Formalin-Fixed, Paraffin-Embedded (FFPE) Hematoxylin and Eosin (H\&E) stained WSIs. These WSIs generate 144,450 image-label pairs across 16 training tasks, with each WSI contributing multiple labels corresponding to different prediction objectives including cancer subtyping, tissue classification, and biomarker prediction.

\subsection{Baselines}
We selected a diverse set of foundation models as baselines to cover both slide-level and patch-level approaches to slide-level classification.
For slide-level models, we included TITAN \citep{titan}, PRISM \citep{prism}, CHIEF \citep{chief}, and Prov-GigaPath \citep{gigapath}, which generate slide-level representations that can be directly used for downstream tasks.
In addition, we incorporated EXAONE Path 1.0 \citep{exaonepath1_0} and UNI2-h \citep{uni} as patch-level foundation model baselines. Although these models operate on localized regions of the slide, their design and prior applications align naturally with slide-level prediction tasks when combined with an appropriate aggregation strategy.
In our experiments, we employed a CLAM-based aggregator \citep{clam} to their patch-level features to produce slide-level predictions.

\subsection{Evaluation Protocols}
Each model was fine-tuned for slide-level classification according to its architectural design, while keeping the pretrained foundation model parameters fixed.
For slide-level foundation models, we trained a linear classification layer on top of the slide-level representations generated by the frozen backbone.
For patch-level foundation models, we adopted the approach proposed in UNI, applying a CLAM aggregator to the patch-level features to generate slide-level predictions.
Our proposed model similarly utilizes patch-level features extracted from the first-stage model, which are then aggregated via CLAM for slide-level inference.
Each benchmark task was evaluated on a predefined training/test split, and we report the average performance over four independent training runs with different random seeds.

\subsection{Slide-Level Benchmarks}
To compare model performance, we construct a total of 10 slide-level benchmark tasks derived from diverse cancer lesions including lung adenocarcinoma, breast cancer, colorectal cancer, and renal cancer.
These benchmarks consist of 4 tasks from private datasets and 6 tasks from public datasets, carefully selected to evaluate both task diversity and model generalization across different data sources and institutions.

\begin{table}[t]
\centering
\caption{AUROC scores on 10 slide-level tasks}
\label{tab:slide-level-results}
\vspace{0.1cm}

\small
\setlength{\tabcolsep}{3pt} % Reduce column padding
\resizebox{\linewidth}{!}{%
\begin{tabular}{lccccccc}
\toprule
\textbf{Benchmarks} & \textbf{TITAN} & \textbf{PRISM} & \textbf{CHIEF} & \textbf{Prov-GigaPath} & \textbf{UNI2-h} & \textbf{EXAONE Path 1.0} & \textbf{EXAONE Path 2.0} \\
\midrule
LUAD-TMB-USA1 & 0.690 & 0.645 & 0.650 & 0.674 & 0.669 & 0.692 & 0.664 \\
%LUAD-TMB-UPMC & 0.6905 & 0.6822 & 0.6460 & 0.6721 & 0.6861 & 0.6677 & 0.6617 \\
LUAD-EGFR-USA1 & 0.754 & 0.815 & 0.784 & 0.709 & 0.827 & 0.784 & 0.853 \\
%LUAD-EGFR-UPMC & 0.8197 & 0.8152 & 0.7691 & 0.7623 & 0.8577 & 0.8488 & 0.7727 \\
%LUAD-KRAS-in\_house & 0.5705 & 0.5494 & 0.5384 & 0.6058 & 0.6158 & 0.5401 & 0.5964 \\
LUAD-KRAS-USA2 & 0.541 & 0.623 & 0.468 & 0.511 & 0.469 & 0.527 & 0.645 \\
CRC-MSI-KOR & 0.937 & 0.943 & 0.927 & 0.954 & 0.981 & 0.972 & 0.938 \\
%BRCA-ER-BCNB & 0.9343 & 0.8998 & 0.9115 & 0.9186 & 0.9454 & 0.9195 & 0.9195 \\
%BRCA-PR-BCNB & 0.8804 & 0.8613 & 0.8470 & 0.8595 & 0.8770 & 0.8406 & 0.8406 \\
%BRCA-HER2-BCNB & 0.8046 & 0.8154 & 0.7822 & 0.7891 & 0.8322 & 0.8212 & 0.8212 \\
BRCA-TP53-CPTAC & 0.788 & 0.842 & 0.788 & 0.739 & 0.808 & 0.766 & 0.757 \\
BRCA-PIK3CA-CPTAC & 0.758 & 0.893 & 0.702 & 0.735 & 0.857 & 0.735 & 0.804 \\
RCC-PBRM1-CPTAC & 0.638 & 0.557 & 0.513 & 0.527 & 0.501 & 0.526 & 0.583 \\
RCC-BAP1-CPTAC & 0.719 & 0.769 & 0.731 & 0.697 & 0.716 & 0.719 & 0.807 \\
COAD-KRAS-CPTAC & 0.764 & 0.744 & 0.699 & 0.815 & 0.943 & 0.767 & 0.912 \\
COAD-TP53-CPTAC & 0.889 & 0.816 & 0.701 & 0.712 & 0.783 & 0.819 & 0.875 \\
\midrule
\textbf{Average} & 0.748 & 0.765 & 0.696 & 0.707 & 0.755 & 0.731 & \textbf{0.784} \\
%\textbf{Average} & 0.761 & 0.767 & 0.716 & 0.730 & 0.773 & 0.747 & \textbf{0.778} \\
\bottomrule
\end{tabular}%
}
\end{table}

\subsubsection{Benchmarks from Private Datasets}
%해당 벤치마크는 Samsung Medical Center (SMC) 및 **University of Pittsburgh Medical Center (UPMC)**와의 협력을 통해 수집된 내부 데이터를 기반으로 구성되었습니다. 모든 데이터 활용은 연구 목적으로 각 기관의 IRB(임상윤리심의위원회) 승인을 받았으며, in-house 검증 데이터는 비식별화 처리된 잠금 데이터로, 외부 활용 없이 내부 성능 검증 용도로만 사용되었습니다.(정정)
%These benchmarks are based on internal datasets collected in collaboration with Samsung Medical Center (SMC) and University of Pittsburgh Medical Center (UPMC). All data usage was approved by the respective Institutional Review Boards (IRBs) for research purposes. The in-house validation data are de-identified and locked only for internal use, and were used strictly for internal performance evaluation.
These benchmarks are based on internal datasets collected in collaboration with one general hospital from Korea (KOR) and two general hospitals from USA (USA1, USA2). All data usage has been approved by the respective Institutional Review Boards (IRBs) for research purposes. All data are de-identified and locked only for internal use, and were used strictly for internal performance evaluation.

\textbf{LUAD-TMB.} This task predicts tumor mutation burden (TMB) status (high vs. low) from lung adenocarcinoma WSIs. TMB is defined as the number of mutations per megabase in DNA sequencing, with a threshold of 10 used to distinguish between high and low. Models were trained on KOR-LUAD (low:high = 1063:287), and tested on USA1-LUAD (137:117) datasets.

\textbf{LUAD-EGFR.} This task detects the presence of EGFR mutations in lung adenocarcinoma. Clinically, mutations of second tier or higher are labeled as "mutated", and all others as "wild type". Training used KOR-LUAD (wild:mut = 1145:205), with testing on USA1-LUAD (242:12).

\textbf{LUAD-KRAS.} This task identifies KRAS mutations in lung adenocarcinoma WSIs using the same clinical mutation criteria as EGFR. Training used KOR1-LUAD (wild:mut = 1217:133), with testing on USA2-LUAD (347:168).

\textbf{CRC-MSI.} This task classifies microsatellite instability (MSI) status in colorectal adenocarcinoma. Models were trained on KOR-CRC (stable:instable = 2630:831) and tested on a held-out portion of the same dataset (658:209).

\subsubsection{Benchmarks from Public Datasets}
%These benchmarks are constructed using publicly available datasets, including BCNB\citep{bcnb} and CPTAC\citep{cptac}, which are widely used in computational pathology research.
These benchmarks are constructed using publicly available dataset, CPTAC \citep{cptac}, which is widely used in computational pathology research.

\textbf{BRCA-TP53, PIK3CA.} These tasks predict TP53 and PIK3CA mutation status from breast cancer WSIs. Both tasks use the CPTAC-BRCA \citep{cptac-brca} dataset with TP53 having train (wild:mut = 53:37), test (14:8) and PIK3CA having train (58:33), test (14:7).

\textbf{RCC-PBRM1, BAP1.} These tasks focus on detecting PBRM1 and BAP1 mutations in clear cell renal cell carcinoma (CCRCC). Both benchmarks use the CPTAC-CCRCC \citep{cptac-ccrcc} dataset with PBRM1 having train (wild:mut = 97:96), test (26:26) and BAP1 having train (156:39), test (46:4).

\textbf{COAD-KRAS, TP53.} These tasks classify KRAS and TP53 mutation status in colon adenocarcinoma. Both use the CPTAC-COAD \citep{cptac-coad} dataset with KRAS having train (wild:mut = 50:29), test (11:8) and TP53 having train (53:27), test (12:6)

\begin{figure}[t]
    \centering
    \includegraphics[width=\linewidth]{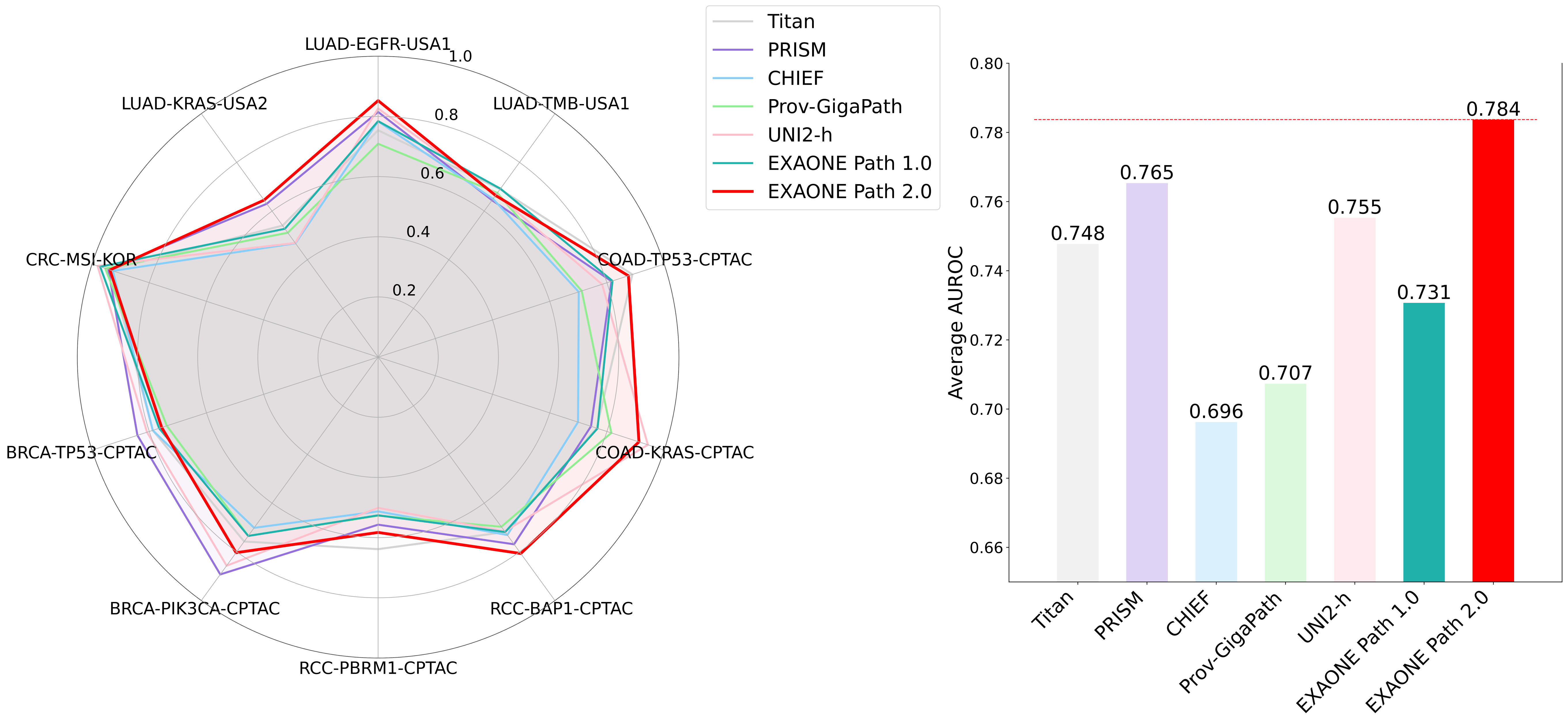}
    \caption{Comparison of AUROC scores across 10 slide-level benchmarks and their averages. EXAONE Path 2.0 (red) shows the most balanced and high-performing profile.}
    \label{fig:radar}
\end{figure}

\subsection{Evaluation Results}
%표~\ref{tab:slide-level-results}는 총 16개의 슬라이드 수준 벤치마크 과제에 대한 7개 모델의 성능을 비교한 결과를 제시합니다. 평가된 모든 모델 중에서, 본 연구에서 제안한 EXAONEPath V2.0이 가장 높은 평균 성능을 기록하였으며, 다양한 조직 유형, 기관, 예측 과제에 걸쳐 안정적인 정확도와 일반화 능력을 보여주었습니다.
Table~\ref{tab:slide-level-results} presents the comparative performance of seven models across 10 slide-level benchmark tasks. Among all evaluated models, EXAONE Path 2.0 achieved the highest overall average performance, demonstrating both robust accuracy and consistent generalization across diverse tissue types, institutions, and prediction targets.

%폐선암(lung adenocarcinoma) 관련 분자 예측 과제에서는 EXAONEPath V2.0이 두드러진 성능을 보였습니다. 특히 EGFR 변이 예측 과제에서는 in-house 데이터셋에서 **가장 높은 정확도(0.8526)**를 기록하였고, UPMC 데이터셋에서도 경쟁력 있는 성능을 보였습니다. KRAS 변이 예측에서는 UPMC 데이터셋에서 **모든 모델 중 최고 성능(0.6451)**을 달성하였으며, in-house 데이터셋에서도 Prov-GigaPath 다음으로 높은 수치를 보였습니다. TMB 분류에서는 TITAN이나 EXAONEPath V1.0보다는 다소 낮았지만, 여전히 상위권의 성능을 유지하였습니다.
In lung adenocarcinoma-related tasks, EXAONE Path 2.0 showed outstanding performance in EGFR mutation prediction, achieving the highest accuracy (0.853) on the USA1-LUAD dataset. In the KRAS mutation task, the model recorded the best performance (0.645) on the USA2-LUAD dataset, surpassing all other baselines. For TMB classification, EXAONE Path 2.0 performed comparably to the top-performing models, although slightly behind EXAONE Path 1.0 and TITAN.

%대장암(colorectal cancer)의 MSI 분류 과제에서는 EXAONEPath V2.0이 0.9377의 높은 정확도를 보였으며, 다른 파운데이션 모델들과 유사한 수준의 견고한 성능을 나타냈습니다.
In colorectal cancer MSI classification, EXAONE Path 2.0 maintained high accuracy (0.938), on par with other foundation models, and showed stable generalization across test sets.

%유방암(breast cancer) 관련 과제에서도 EXAONEPath V2.0은 ER, PR, HER2 리셉터 과제와 TP53, PIK3CA 유전자 변이 예측 과제 모두에서 상위권 성능을 지속적으로 유지하였습니다. 최상위 성능을 기록한 경우는 적지만, 모든 과제에서 상위 2~3위권 내에 꾸준히 포지셔닝되며 안정적인 결과를 보여주었습니다.
In breast cancer tasks, the model consistently produced strong results across all  mutation (TP53, PIK3CA) benchmarks. While it did not always achieve the highest score, it demonstrated reliable performance even in challenging classification scenarios with limited training samples.

%신장암(renal cell carcinoma) 과제에서는 특히 **BAP1 변이 예측에서 가장 높은 성능(0.8070)**을 달성하였고, PBRM1 예측 과제에서도 TITAN 다음으로 높은 성능을 기록하며 경쟁력을 입증했습니다. 또한, 결장암(colon adenocarcinoma) 관련 KRAS 및 TP53 예측 과제에서도 각각 0.9119와 0.8750의 높은 정확도를 기록하며 UNI2-h 및 TITAN에 필적하는 수준의 성능을 보였습니다.
In the RCC benchmarks, EXAONE Path 2.0 showed clear superiority in the BAP1 mutation task, achieving the highest score (0.807), and performed competitively in the PBRM1 benchmark as well. In the colon adenocarcinoma benchmarks, the model reached top-tier results, including a near-optimal score of 0.912 in KRAS prediction and 0.875 in TP53 mutation classification.

%총 16개 벤치마크 중 5개 과제에서 최고 성능을 기록하였으며, 대부분의 과제에서 상위 3위 이내에 들었습니다. 이러한 결과는 EXAONEPath V2.0의 계층적 구조 설계와 슬라이드 수준 라벨만을 활용한 end-to-end 학습 전략이 실제 병리학적 다운스트림 작업에서 실질적인 성능 향상을 이끌어낼 수 있음을 실증적으로 보여줍니다. EXAONEPath V2.0은 다양한 병리학 과제 전반에서 새로운 기준을 제시하는 파운데이션 모델로 자리매김하였습니다.
Overall, EXAONE Path 2.0 achieved the best average AUROC score and remained within the top three across nearly all tasks. These results empirically validate the benefits of our unified hierarchical framework and end-to-end optimization strategy, showing that EXAONE Path 2.0 can serve as a strong and generalizable foundation model for a wide range of slide-level pathology tasks.

%모든 벤치마크에 대한 포괄적인 비교를 제공하기 위해, 우리는 모델 성능을 레이더 다이어그램(Figure~\ref{fig:radar})으로 시각화하였다. 이 스파이더 플롯은 16개의 검증 데이터셋에 대한 각 모델의 정규화된 점수를 보여주며, 성능의 일관성을 직관적으로 이해할 수 있도록 한다. 도표에서 확인할 수 있듯이, EXAONE Path V2.0은 모든 벤치마크 축에서 안정적으로 높은 성능을 보여주며, 균형 잡히고 넓은 면적을 형성한다. 이는 EXAONE Path V2.0이 다른 파운데이션 모델들보다 뛰어난 일반화 성능과 강건함을 갖추고 있음을 시사한다. 많은 경쟁 모델들이 특정 과제에서 성능 저하를 보이는 반면, EXAONE Path V2.0은 시각적으로도 두드러진 프로파일을 보여주며, 슬라이드 레벨 파운데이션 모델로서의 적합성을 뒷받침한다.
To provide a holistic comparison across all benchmarks, we visualize model performance using radar and bar charts (Figure~\ref{fig:radar}). The charts illustrate the AUROC of each model across 10 validation datasets, enabling an intuitive understanding of performance consistency. As shown, EXAONE Path 2.0 demonstrates consistently strong coverage across all benchmarks. This indicates its superior generalization capability and robustness compared to other foundation models, many of which exhibit performance dips on specific tasks. The visually dominant profile of EXAONE Path 2.0 reinforces its leading average performance and highlights its suitability as a universal slide-level foundation model.

% \section{Results}
% \input{5.Results}

% \section{Discussion}
% \input{6.discussion}

\section{Conclusion}
We presented EXAONE Path 2.0, a pathology foundation model that learns patch-level representations under direct slide-level supervision. Our approach enables slide-level supervised signals to propagate through all hierarchical stages, allowing end-to-end learning of clinically relevant representations.

Our method addresses computational challenges through hierarchical architecture design, curriculum learning, and memory management techniques including activation checkpointing and CPU offloading. We employ multi-task learning across diverse biomarker prediction tasks and use early exit strategies to mitigate overfitting in small data regimes.

Experimental results show that EXAONE Path 2.0 achieves competitive average performance across 10 biomarker prediction tasks using only 37k WSIs for training, demonstrating improved data efficiency compared to existing foundation models. The model performs consistently across diverse cancer types and prediction targets.

These results demonstrate that direct slide-level supervision can effectively learn clinically relevant features, and our proposed methods successfully address the computational challenges of gigapixel image training, providing a practical approach for pathology foundation models.

% Bibliography
\bibliography{8.reference}

\newpage
\section*{EXAONEPath AI Model License Agreement 1.0 - NC}
This License Agreement (“Agreement”) is entered into between you (“Licensee”) and LG Management Development Institute Co., Ltd. (“Licensor”), governing the use of the EXAONEPath AI Model (“Model”). By downloading, installing, copying, or using the Model, you agree to comply with and be bound by the terms of this Agreement. If you do not agree to all the terms, you must not download, install, copy, or use the Model. This Agreement constitutes a binding legal agreement between the Licensee and Licensor.
\section*{1. Definitions}
\subsection*{1.1 Model}The artificial intelligence model provided by Licensor, which includes any software, algorithms, machine learning models, or related components supplied by Licensor. This definition extends to encompass all updates, enhancements, improvements, bug fixes, patches, or other modifications that may be provided by Licensor from time to time, whether automatically or manually implemented.
\subsection*{1.2 Derivatives}Any modifications, alterations, enhancements, improvements, adaptations, or derivative works of the Model created by Licensee or any third party. This includes changes made to the Model's architecture, parameters, data processing methods, or any other aspect of the Model that results in a modification of its functionality or output.
\subsection*{1.3 Output} Any data, results, content, predictions, analyses, insights, or other materials generated by the Model or Derivatives, regardless of whether they are in their original form or have been further processed or modified by the Licensee. This includes, but is not limited to, textual or numerical produced directly or indirectly through the use of the Model.
\subsection*{1.4 Licensor} LG Management Development Institute Co., Ltd., the owner, developer, and provider of the EXAONEPath AI Model. The Licensor holds all rights, title, and interest in the Model and is responsible for granting licenses to use the Model under the terms specified in this Agreement.
\subsection*{1.5 Licensee} The individual, organization, corporation, academic institution, government agency, or other entity using or intending to use the Model under the terms and conditions of this Agreement. The Licensee is responsible for ensuring compliance with the Agreement by all authorized users who access or utilize the Model on behalf of the Licensee.
\section*{2. License Grant}
\subsection*{2.1 Grant of License} Subject to the terms and conditions outlined in this Agreement, the Licensor hereby grants the Licensee a limited, non-exclusive, non-transferable, worldwide, and revocable license to:
\begin{enumerate}[label=\alph*.]
    \item Access, download, install, and use the Model solely for research purposes. This includes evaluation, testing, academic research and experimentation.
    \item Publicly disclose research results and findings derived from the use of the Model or Derivatives, including publishing papers or presentations.
    \item Modify the Model and create Derivatives based on the Model, provided that such modifications and Derivatives are used exclusively for research purposes. The Licensee may conduct experiments, perform analyses, and apply custom modifications to the Model to explore its capabilities and performance under various scenarios. If the Model is modified, the modified Model must include "EXAONEPath" at the beginning of its name.
    \item Distribute the Model and Derivatives in each case with a copy of this Agreement.
\end{enumerate}
\subsection*{2.2 Scope of License} The license granted herein does not authorize the Licensee to use the Model for any purpose not explicitly permitted under this Agreement. Any use beyond the scope of this license, including any commercial application or external distribution, is strictly prohibited unless explicitly agreed upon in writing by the Licensor.
\section*{3. Restrictions}
\subsection*{3.1 Commercial Use} The Licensee is expressly prohibited from using the Model, Derivatives, or Output for any commercial purposes, including but not limited to, developing or deploying products, services, or applications that generate revenue, whether directly or indirectly. Any commercial exploitation of the Model or its derivatives requires a separate commercial license agreement with the Licensor. Furthermore, the Licensee shall not use the Model, Derivatives or Output to develop or improve other models, except for research purposes, which is explicitly permitted.
\subsection*{3.2 Reverse Engineering} The Licensee shall not decompile, disassemble, reverse engineer, or attempt to derive the source code, underlying ideas, algorithms, or structure of the Model, except to the extent that such activities are expressly permitted by applicable law. Any attempt to bypass or circumvent technological protection measures applied to the Model is strictly prohibited.
\subsection*{3.3 Unlawful Use} The Licensee shall not use the Model and Derivatives for any illegal, fraudulent, or unauthorized activities, nor for any purpose that violates applicable laws or regulations. This includes but is not limited to the creation, distribution, or dissemination of malicious, deceptive, or unlawful content.
\subsection*{3.4 Ethical Use} The Licensee shall ensure that the Model or Derivatives is used in an ethical and responsible manner, adhering to the following guidelines:
\begin{enumerate}[label=\alph*.]
    \item The Model and Derivatives shall not be used to generate, propagate, or amplify false, misleading, or harmful information, including fake news, misinformation, or disinformation.
    \item The Model and Derivatives shall not be employed to create, distribute, or promote content that is discriminatory, harassing, defamatory, abusive, or otherwise offensive to individuals or groups based on race, gender, sexual orientation, religion, nationality, or other protected characteristics.
    \item The Model and Derivatives shall not infringe on the rights of others, including intellectual property rights, privacy rights, or any other rights recognized by law. The Licensee shall obtain all necessary permissions and consents before using the Model and Derivatives in a manner that may impact the rights of third parties.
    \item The Model and Derivatives shall not be used in a way that causes harm, whether physical, mental, emotional, or financial, to individuals, organizations, or communities. The Licensee shall take all reasonable measures to prevent misuse or abuse of the Model and Derivatives that could result in harm or injury.
\end{enumerate}

\section*{4. Ownership}
\subsection*{4.1 Intellectual Property} All rights, title, and interest in and to the Model, including any modifications, Derivatives, and associated documentation, are and shall remain the exclusive property of the Licensor. The Licensee acknowledges that this Agreement does not transfer any ownership rights to the Licensee. All trademarks, service marks, and logos associated with the Model are the property of the Licensor.
\subsection*{4.2 Output} All output generated by the Model from Licensee Data ("Output") shall be the sole property of the Licensee. Licensor hereby waives any claim of ownership or intellectual property rights to the Output. Licensee is solely responsible for the legality, accuracy, quality, integrity, and use of the Output.
\subsection*{4.3 Attribution} In any publication or presentation of results obtained using the Model, the Licensee shall provide appropriate attribution to the Licensor, citing the Model's name and version, along with any relevant documentation or references specified by the Licensor.
\section*{5. No Warranty}
\subsection*{5.1 “As-Is” Basis} The Model, Derivatives, and Output are provided on an “as-is” and “as-available” basis, without any warranties or representations of any kind, whether express, implied, or statutory. The Licensor disclaims all warranties, including but not limited to, implied warranties of merchantability, fitness for a particular purpose, accuracy, reliability, non-infringement, or any warranty arising from the course of dealing or usage of trade.
\subsection*{5.2 Performance and Reliability} The Licensor does not warrant or guarantee that the Model, Derivatives or Output will meet the Licensee’s requirements, that the operation of the Model, Derivatives or Output will be uninterrupted or error-free, or that defects in the Model will be corrected. The Licensee acknowledges that the use of the Model, Derivatives or Output is at its own risk and that the Model, Derivatives or Output may contain bugs, errors, or other limitations.
\subsection*{5.3 No Endorsement} The Licensor does not endorse, approve, or certify any results, conclusions, or recommendations derived from the use of the Model. The Licensee is solely responsible for evaluating the accuracy, reliability, and suitability of the Model for its intended purposes.
\section*{6. Limitation of Liability}
\subsection*{6.1 No Liability for Damages} To the fullest extent permitted by applicable law, in no event shall the Licensor be liable for any special, incidental, indirect, consequential, exemplary, or punitive damages, including but not limited to, damages for loss of business profits, business interruption, loss of business information, loss of data, or any other pecuniary or non-pecuniary loss arising out of or in connection with the use or inability to use the Model, Derivatives or any Output, even if the Licensor has been advised of the possibility of such damages.
\subsection*{6.2 Indemnification} The Licensee agrees to indemnify, defend, and hold harmless the Licensor, its affiliates, officers, directors, employees, and agents from and against any claims, liabilities, damages, losses, costs, or expenses (including reasonable attorneys' fees) arising out of or related to the Licensee's use of the Model, any Derivatives, or any Output, including any violation of this Agreement or applicable laws. This includes, but is not limited to, ensuring compliance with copyright laws, privacy regulations, defamation laws, and any other applicable legal or regulatory requirements.
\section*{7. Termination}
\subsection*{7.1 Termination by Licensor} The Licensor reserves the right to terminate this Agreement and revoke the Licensee’s rights to use the Model at any time, with or without cause, and without prior notice if the Licensee breaches any of the terms or conditions of this Agreement. Termination shall be effective immediately upon notice.
\subsection*{7.2 Effect of Termination} Upon termination of this Agreement, the Licensee must immediately cease all use of the Model, Derivatives, and Output and destroy all copies of the Model, Derivatives, and Output in its possession or control, including any backup or archival copies. The Licensee shall certify in writing to the Licensor that such destruction has been completed.
\subsection*{7.3 Survival} The provisions of this Agreement that by their nature should survive termination, including but not limited to, Sections 4 (Ownership), 5 (No Warranty), 6 (Limitation of Liability), and this Section 7 (Termination), shall continue to apply after termination.
\section*{8. Governing Law}
\subsection*{8.1 Governing Law} This Agreement shall be governed by and construed in accordance with the laws of the Republic of Korea, without regard to its conflict of laws principles.
\subsection*{8.2 Arbitration} Any disputes, controversies, or claims arising out of or relating to this Agreement, including its existence, validity, interpretation, performance, breach, or termination, shall be referred to and finally resolved by arbitration administered by the Korean Commercial Arbitration Board (KCAB) in accordance with the International Arbitration Rules of the Korean Commercial Arbitration Board in force at the time of the commencement of the arbitration. The seat of arbitration shall be Seoul, Republic of Korea. The tribunal shall consist of one arbitrator. The language of the arbitration shall be English.
\section*{9. Alterations}
\subsection*{9.1 Modifications} The Licensor reserves the right to modify or amend this Agreement at any time, in its sole discretion. Any modifications will be effective upon posting the updated Agreement on the Licensor’s website or through other means of communication. The Licensee is responsible for reviewing the Agreement periodically for changes. Continued use of the Model after any modifications have been made constitutes acceptance of the revised Agreement.
\subsection*{9.2 Entire Agreement} This Agreement constitutes the entire agreement between the Licensee and Licensor concerning the subject matter hereof and supersedes all prior or contemporaneous oral or written agreements, representations, or understandings. Any terms or conditions of any purchase order or other document submitted by the Licensee in connection with the Model that are in addition to, different from, or inconsistent with the terms and conditions of this Agreement are not binding on the Licensor and are void.\\

By downloading, installing, or using the EXAONEPath AI Model, the Licensee acknowledges that it has read, understood, and agrees to be bound by the terms and conditions of this Agreement.

\end{document}